\documentclass[cameraready]{Interspeech}

\usepackage{graphicx}

\title{Umm... With Transformers? \\Insights from Filled Pause Use across Four Slavic Parliaments}

\author[affiliation={1}, orcid=0009-0005-8013-8257, correspondingauthor]{Ivan}{Porupski}
\author[affiliation={2, 3}, orcid=0000-0001-5748-2643]{Branimir}{Dropuljić}
\author[affiliation={1,4,5}, orcid=0000-0001-7169-9152]{Nikola}{Ljubešić}

\address{
    $^1$ Department of Knowledge Technologies, Jožef Stefan Institute, Ljubljana, Slovenia \\
    $^2$ TransUnion, Zagreb, Croatia \\
    $^3$ Faculty of Electrical Engineering and Computing, University of Zagreb, Croatia \\
    $^4$ Faculty of Computer and Information Science, University of Ljubljana, Slovenia \\
    $^5$ Institute of Contemporary History, Ljubljana, Slovenia
}

\email{ivan.porupski@ijs.si, branimir.dropuljic@fer.hr, nikola.ljubesic@ijs.si}


\begin{document}

\maketitle

\begin{abstract}
  Filled pauses (FPs) are a universal feature of spontaneous speech, yet most studies rely on small, single-language corpora, limiting the generalisability of their findings. We analyse $\sim$4,000 hours of parliamentary speech across four related Slavic languages (Croatian, Czech, Polish, Serbian). FP occurrence is obtained via transformer-based automatic detection, while FP rate is modelled using Generalised Estimating Equations (GEE) with Mundlak correction to distinguish within- from between-speaker effects. We replicate a negative association of age and speech rate with FP rate, but find that gender effects are language-specific and directionally opposite to most prior literature. Novel analyses of sentiment, political orientation, and power status reveal a consistent positive association between sentiment and FP rate, alongside parliament-specific modulation by orientation and power status, with opposition speakers tending toward lower FP rates than governing coalition speakers.
\end{abstract}

\section{Introduction}

Filled pauses --- vocalisations such as \textit{uh} and \textit{um} --- are a pervasive feature of spontaneous speech, serving functions ranging from turn-holding to signalling lexical difficulty~\cite{clark2002using}. Despite sustained research interest, most empirical work relies on small, single-language corpora, which limits both statistical power and the generalisability of conclusions across linguistic and cultural contexts.

We address this gap by analysing FP production across almost 4,000 hours of parliamentary speech from four related Slavic languages, using a transformer-based detector for automatic FP identification at scale. Parliamentary speech offers a controlled domain with rich speaker metadata, enabling investigation of variables rarely studied together: gender, age, speech rate, expressed sentiment, political orientation, and power status.
Our analyses are partly confirmatory --- we revisit established predictors (gender, age, speech rate) in a larger and more diverse setting than typical --- and partly exploratory, examining sentiment, political orientation, and power status as novel predictors of FP rate. 
Crucially, we apply Mundlak-corrected GEE models to decompose predictors into stable between-speaker tendencies and utterance-level within-speaker variation, allowing us to distinguish trait-level from state-level associations.

\section{Related Work}


\subsection{Automatic FP Identification} 
Early filled-pause detection relied on hand-crafted acoustic cues (pitch/F0, MFCCs, formants, spectral/vocal-tract stability) with modest precision and recall~\cite{goto1999real,stouten2003feature,audhkhasi2009formant}. Prosodic-feature methods improved results (0.61 F1), and prosodic discontinuity features reached 0.83 F1 in spontaneous speech~\cite{medeiros2013experiments,reichel2019filled}. Recent automation increasingly uses speech transformers (e.g., wav2vec2, HuBERT, WavLM), reaching strong frame-level F1 (0.86-0.88) on Switchboard—a prominent general-population speech benchmark~\cite{romana2023automatic,godfrey1992switchboard}. Recently, a study using wav2vec2-bert, Slovenian training data, and Croatian, Czech, Polish and Serbian parliamentary data for testing, showed very strong cross-lingual performance, with event-level F1 performance of 0.87-0.94. This performance significantly surpasses human recall, albeit with a minor hit on precision \cite{ljubevsic2025identifying}. Beyond these two, most newer work targets atypical speech (stuttering) and explores transfer and multitask learning across corpora/languages, but may not generalise to regular speech~\cite{lea2021sep,bayerl2022ksof,romana2024fluencybank,mohapatra2022speech,bayerl2022detecting,liu2023automatic}. 


\subsection{Gender}
In English task-oriented dialogue, men are reported to produce more verbal fillers (including FPs) than women (3.04 vs. 2.07 per 100 words), and the director role amplifies fillers more for men than for women, indicating an interaction between gender and role~\cite{bortfeld2001disfluency}.
Similarly, a study of British and American English corpora found that men not only produce more FPs than women but also show a distinct preference for \emph{uh} over \emph{um}, whereas women tend toward higher relative proportions of \emph{um}~\cite{tottie2011uh_british,tottie2014use_american}. In Dutch spontaneous speech (the CGN corpus), male speakers show higher filled-pause rates than female speakers in both face-to-face and telephone conversations; importantly, this difference is reported as holding regardless of interlocutor gender~\cite{binnenpoorte2005gender}.
In some corpora (including Switchboard), men have been reported to use hesitation markers at higher overall rates, but the most consistent cross-corpus pattern concerns the \emph{um}/\emph{uh} split (women and younger speakers favour \emph{um})~\cite{wieling2016variation}.
This suggests that the communicative context may modulate gendered patterns: gender differences prominent in informal conversation often neutralise in more formal or task-oriented genres~\cite{crible2018discourse}.
In a small, balanced sample of European Parliament speeches, Mikl\'ossiov\'a~\cite{miklossiova2023role} reports minimal gender differences in pause parameters, with FPs relatively rare and frequency differences interpreted cautiously given the limited data.

\subsection{Age}
In Bortfeld et al.'s task-oriented dialogues, age effects were concentrated in the oldest group (63–72) and were smaller in magnitude than the strong effects of task role and domain on filler/disfluency rates, and consistent with an account in which planning demands drive much of the variability~\cite{bortfeld2001disfluency}.
In a large English conversational corpus spanning ages 17--68, however, rate of FPs shows a positive association with age (older speakers producing more FPs) 
along with slower speech rate. Age and rate were not significantly correlated once FP rate was controlled~\cite{horton2010corpus}.
By contrast, a large-scale study suggests that while total FP rates may appear comparable across ages, the composition shifts: \emph{uh} usage increases significantly with age while \emph{um} decreases~\cite{laserna2014}. This trend is further supported by evidence from European Portuguese, where FPs, unlike other disfluency types, show a distinct increase across the lifespan~\cite{moniz2014disfluencies}.
In cross-Germanic corpora, age is especially visible in marker choice (younger speakers leading the shift from \emph{uh} toward \emph{um}) and in a decrease in FP use with age in the Switchboard corpus, while smaller corpora not providing clear trends~\cite{wieling2016variation}. Very recently, a study on English spoken narratives (story re-telling task) found that FP frequency decreases with age -- younger adults produce significantly more total FPs~\cite{engelhardt_markostamou_2025_disfluency}.

\subsection{Speech rate}
In spontaneous speech, FP production covaries with speaking rate: slower speech tends to contain more FPs~\cite{clark2002using}.
Early corpus work already showed strong between-speaker differences, where faster speakers exhibit lower rates of hesitation phenomena (including FPs)~\cite{maclay1959hesitation}.
Large-scale Switchboard analyses likewise find a negative correlation between speech rate and FP rate (slower speakers produce more fillers), reflecting both the delay introduced by FPs themselves and the fact that some speakers remain slow even when FPs are excluded from rate calculations~\cite{horton2010corpus,shriberg2001errrr}.
This relationship is consistent with accounts linking slowed articulation and FPs to increased planning and lexical-retrieval demands~\cite{clark2002using}.

\subsection{Sentiment}
Work relating affective state to FPs mainly implicates expectancy (certainty) and interactional conflict rather than valence polarity. 
Within affective-dialogue research, disfluency features (including FPs) are treated as paralinguistic cues that complement lexical and acoustic information. In dialogue emotion recognition, disfluency features are not strongest on their own, but improve performance when added to lexical/acoustic models~\cite{tian_recognizing_2015}. 
In dimensional settings, these cues are typically more informative for expectancy than for valence (positive-negative affect), where effects are weaker and distributions overlap substantially~\cite{tian_era_2018}.
In Slovenian, FPs are reported as relatively more common in public speech, interpreted as reflecting time pressure and emotional stress~\cite{verdonik2025publicprivate}. Overall, the strongest, most reproducible link with FPs relates to uncertainty or stress rather than sentiment polarity.

\subsection{Political orientation}
Direct empirical evidence linking left–right ideology to filled-pause rates is limited; existing work is largely qualitative and speaker-specific. 
What has been done is mostly case-based analysis of individual politicians' speaking styles rather than systematic ideology comparisons. For example, a detailed analysis of a prominent US politician's campaign speech argues that a conspicuous near-absence of FPs is part of an intentionally ``informal'' populist vocal style~\cite{kjeldgaardchristiansen2024voice}. In contrast, a recent study of a US political interview inventories and functionally classifies fillers, but does not frame results as ideology effects -- it is descriptive of one speaker and setting~\cite{alfaragy2025biden}. 

\subsection{Power status}
Work explicitly relating government–opposition status to filled-pause rates is limited; existing corpus evidence more often addresses prosodic style. For example, Finnish parliamentary speech shows systematic differences in mean F0 
by opposition vs government status, with important corpus-design and metadata caveats~\cite{vainio2023prosody}.
Recent Bundestag analyses likewise report style shifts when parties move between government and opposition, though results are presented at the level of discourse style and sentiment rather than use of FPs~\cite{patz2025bundestag}.

\subsection{Our Contribution}

We revisit established predictors of FP production --- gender, age, and speech rate --- in a substantially larger and more linguistically diverse setting than prior work, testing the generalisability of reported effects across four Slavic parliamentary cultures. 
We further explore three predictors with little or no prior empirical grounding: expressed sentiment, political orientation, and power status.
We therefore treat political orientation as an exploratory factor rather than a well-established predictor of filled-pause rate.
Crucially, we apply within-speaker decomposition to disentangle stable speaker traits from utterance-level variation, allowing us to distinguish trait-level from state-level associations. 
Our data come exclusively from the parliamentary domain, which is an important limitation to keep in mind when interpreting the results.

\section{Data}

We use the ParlaSpeech dataset\footnote{\url{https://clarinsi.github.io/parlaspeech/}}~\cite{ljubevsic2025parlaspeech} of 6,000 hours of parliamentary speech from the Croatian (HR, 2015-2022), Czech (CZ, 2013-2023), Polish (PL, 2017-2022), and Serbian (RS, 2013-2022) parliaments, with aligned official transcripts. Speaker metadata (gender, age, political orientation, party power status) were taken from official parliamentary records.
Utterance-level sentiment was automatically predicted via XLM-R-ParlaSent~\cite{mochtak2024parlasent} with $R^2 \approx 0.65$ performance, comparable to the inter-annotator agreement upper bound; filled-pause occurrence was automatically predicted via wav2vec2-bert with event-level $F1 \approx 0.92$ across all four languages~\cite{ljubevsic2025identifying}, again comparable to the inter-annotator agreement; speech rate was derived as (syllabic) vowel count divided by audio duration.
We restrict the analysis to regular members of parliament, utterances of at least 3 seconds and 10 words, 
and minimally 10 utterances per speaker, yielding 1,001,787 utterances from 1,561 speakers (3,889 hours).

\section{Method}

\subsection{Model}

Having detected FPs automatically, we model their \textit{rate of occurrence} across utterances to quantify how speaker- and utterance-level factors relate to FP production. We treat FP count per utterance as a Negative Binomial outcome, chosen for its robustness to overdispersion typical of count data, with $\log(\text{audio duration})$ as offset to express effects as rates. Utterances are nested within speeches within speakers, making the true correlation structure complex. We 
fit the model via Generalised Estimating Equations (GEE)~\cite{liang1986longitudinal} with speakers as clusters, relying on robust sandwich standard errors which remain valid under any misspecification of the working correlation structure. 
We prefer GEE over Generalised Linear Mixed Models (GLMM) 
as it yields population-average estimates directly interpretable as effects on the typical speaker, which aligns with our inferential goal of characterising general tendencies rather than individual-level variation. 
We fit ten models in total: baseline (no Mundlak correction) and Mundlak-corrected specifications, each estimated globally (pooled across all four parliaments) and separately per parliament, all using a GEE Negative Binomial model with independent working correlation structure.

\subsection{Predictors}

Speech rate is z-score standardised (1\,SD\,$\approx$\,1\,syll/s); age is mean-centred,
with effects per decade; sentiment's regression output ranges 0--5 (0--1 is negative, 2--3 is neutral, 4--5 is positive); orientation ranges $-3$ to $+3$ (left to right), both in raw units.

We estimate two specifications. The \textit{baseline} model includes each predictor as a single term. The \textit{Mundlak-corrected} model~\cite{mundlak1978} decomposes each time-varying predictor $x_{it}$ into its speaker mean $\bar{x}_i$ (between-speaker) and utterance deviation $x_{it}^{\text{dev}} = x_{it} - \bar{x}_i$ (within-speaker), separating stable speaker-level tendencies from utterance-to-utterance variation and reducing omitted variable bias from fixed speaker traits.

\begin{equation*}
\begin{aligned}
\mathrm{FP}_{it} \sim & \mathrm{Gender}_i 
  + \mathrm{Age}_{it}
  + \mathrm{Rate}_{it}^{*} \\
  & + \mathrm{Sent}_{it}^{*} 
  + \mathrm{Orient}_{it}^{*} 
  + \mathrm{Status}_{it}^{*} \\
  & + \mathrm{C(Parliament)}_i 
  + \log T_{it}
\end{aligned}
\end{equation*}

\noindent where $x^{*}$ denotes the non-Mundlak baseline ($x_{it}$) or the Mundlak pair $(\bar{x}_i, x_{it}^{\text{dev}})$.

\section{Results}

\begin{figure*}[t]
    \centering
    \includegraphics[width=\textwidth]{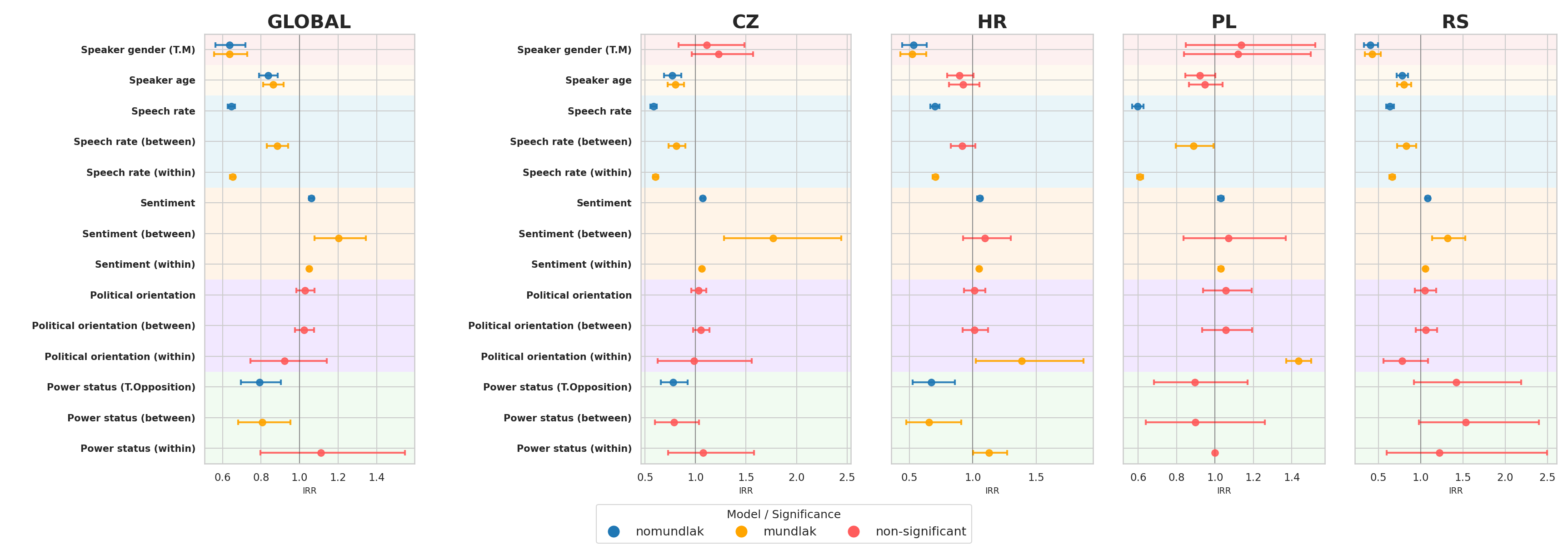}
    \caption{Incidence Rate Ratios (IRR) from GEE Negative Binomial models with independent
    working correlation structure, comparing baseline (no Mundlak correction; blue) and
    Mundlak-corrected (orange) specifications across four parliaments and a pooled global model.
    The Mundlak decomposition partitions time-varying predictors into within-speaker deviation
    and between-speaker mean components. Error bars show 95\% confidence intervals; red markers
    indicate non-significant effects (CI crossing 1.0).}
    \label{fig:irr_results}
\end{figure*}

We first report general incidence of FPs in our data. HR and RS show substantially lower baseline FP rates (1.82 and 1.38 FPs/min, respectively) than CZ and PL (2.91 and 3.47 FPs/min), motivating the inclusion of parliament as a covariate in the global model and suggesting cross-linguistic differences in disfluency norms or parliamentary speech style.
The Incidence Rate Ratio (IRR) 
results are shown in Figure~\ref{fig:irr_results}, with a governing female, Czech speaker of average age (50) with centrist political orientation and negative (0) sentiment as the reference.

\subsection{Gender}

Globally, male speakers produce significantly fewer FPs than female speakers (IRR~$= 0.636$; $-36.4$\% drop in FPs for male speakers), and this effect is stable across both specifications, indicating robustness to the inclusion of between-speaker controls for time-varying covariates.
The effect is entirely driven by HR (IRR~$\approx 0.52$--$0.53$; $\approx -47$\%) and RS (IRR~$\approx 0.40$--$0.42$; $\approx -59$\%), where it is large and consistent; in CZ and PL, gender is non-significant in both specifications, suggesting no reliable gender difference in FP production in those parliaments.
This finding contrasts with prior literature reporting higher FP rates in men~\cite{bortfeld2001disfluency,tottie2011uh_british,binnenpoorte2005gender}. The reversal in direction relative to these predominantly conversational corpora suggests that the gender gap is not universal and may depend on discourse domain and sociolinguistic context.
The higher FP rate in women appears specific to South Slavic parliaments, which calls for further culture-specific focus in future work.

\subsection{Age}

Globally, each additional decade of age is associated with a significant reduction in FPs (IRR~$= 0.862$; $-13.8$\% drop in FPs per decade).
CZ and RS replicate this clearly (IRR~$\approx 0.77$--$0.80$; $\approx -20$\% per decade); HR and PL show the same directional trend but do not reach significance in the between-speaker component (HR: IRR~$= 0.925$, $p = .232$; PL: IRR~$= 0.948$, $p = .259$), possibly reflecting lower between-speaker age variation in those samples. 
Taken together, the results indicate that age-related reductions in FP production are robust in parliamentary speech, but diverge from conversational corpora reporting increases with age, reinforcing the domain-dependence of age effects in disfluency research.

\subsection{Speech rate}

Globally, higher speech rate predicts fewer FPs (IRR~$= 0.645$; $-35.5$\% per\,syll/s). The within-speaker component is the strongest predictor in the dataset (IRR~$= 0.652$; $-34.8$\% per\,syll/s above own baseline), while the between-speaker component is more modest (IRR~$= 0.883$; $-11.7$\% per\,syll/s) and non-significant in HR ($p = .109$), suggesting that in Croatian the cross-speaker association is absorbed by stable speaker-level characteristics. The large within-to-between contrast is consistent with accounts linking FP production to moment-to-moment planning demands rather than a speaker's
habitual pace~\cite{clark2002using,maclay1959hesitation}.

\subsection{Sentiment}

Globally, higher sentiment predicts more FPs (IRR~$= 1.060$; $+6.0$\% per sentiment point),
and this is significant across all four parliaments in the baseline model. Between speakers, those who habitually use more positive language produce more FPs on average (IRR~$= 1.203$; $+20.3$\% per sentiment point), this being significant globally and in CZ and RS, but not in HR ($p = .295$) or PL ($p = .590$). Within speakers, when particular speakers speak more positively, the effect is smaller but consistent across all four parliaments (IRR~$\approx 1.03$--$1.06$; $+3$--$6$\% per point above own baseline). 
While prior work more consistently links FPs to uncertainty or arousal than to valence polarity, the present results suggest a modest but consistent within-speaker association between sentiment and FP rate. This may reflect context-specific production demands in parliamentary speech, though the underlying mechanism remains to be established.

\subsection{Political orientation}

No reliable association between a speaker's habitual ideological position and FP production emerges globally or in any individual parliament (between-speaker: IRR~$= 1.024$, $p = .340$). The within-speaker component, however, is significantly positive in HR (IRR~$= 1.386$; $+38.6$\% per orientation point on the $-3$ to $+3$ scale,
$p = .034$) and PL (IRR~$= 1.434$; $+43.4$\%, $p < .001$), meaning sessions in which a speaker's party affiliation was more rightward than in other sessions are associated with more FPs; 
this effect is absent in CZ and RS, and is masked in the global model (IRR~$= 0.923$, $p = .458$) due to cancellation across language groups. 
The within-speaker signal is confined to HR and PL and absent in CZ, RS, and the global model, indicating that the overall evidence is limited. The effect reflects relatively infrequent episodes in which speakers switch parties that are of different political orientation, and should therefore be interpreted cautiously and explored in future work.

\subsection{Power status}

Globally, opposing-party status is associated with fewer FPs (IRR~$= 0.792$;
$-20.8$\% drop in FPs among opposition members of parliament relative to ruling-coalition members), with significance in CZ (IRR~$= 0.777$) and HR (IRR~$= 0.672$), but not in PL and RS. Serbia shows a borderline positive trend (IRR~$= 1.534$, $p = .061$) -- a directional reversal of the general trend, and may reflect the particular political dynamics of the Serbian parliament during the studied period, which warrants separate
investigation.

Between speakers, the effect is globally significant (IRR~$= 0.805$, $p = .011$) and only in HR (IRR~$= 0.657$; $-34.3$\%, $p = .011$) while attenuates elsewhere. 
Within-speaker variation, i.e., comparing a politician's behaviour while their party is ruling vs. in opposition, is non-significant globally and in CZ, PL, and RS, reaching only borderline significance in HR (IRR~$= 1.129$; $+12.9$\%, $p = .044$).

\section{Conclusion}

We presented a large-scale cross-linguistic analysis of filled-pause production across almost 4,000 hours of parliamentary speech in four Slavic languages, combining transformer-based automatic FP detection with Mundlak-corrected GEE modelling to separate stable speaker traits from utterance-level variation.

For established predictors, we replicate a negative effect of age and confirm the well-documented inverse relationship between speech rate and FP rate. Notably, the within-speaker component is substantially larger than the between-speaker one, suggesting that FPs primarily reflect moment-to-moment planning demands rather than habitual speaking style.

Gender effects appear to be language-specific. The male-higher pattern commonly reported in prior literature reverses in South Slavic parliaments and disappears entirely in West Slavic ones, cautioning against cross-linguistic generalisation and suggesting potential culture-dependent effects.

For novel predictors, sentiment shows a consistent within-speaker effect, meaning speakers use more FPs than usual when they are more positive than their personal baseline. The between-speaker effect is less stable, showing larger effects in some parliaments (CZ and RS) and no significant effects in others (HR and PL). On the global scale, speakers who are generally more positive across their speeches 
exhibit significantly higher FP rates than their more negative colleagues. 
This suggests that the sentiment–FP link is driven more by a speaker’s overall communicative style than by the specific demands of individual speeches. At the same time, the within-speaker effect appears more stable across parliaments.
Political orientation and power status yielded more fragmented patterns, with parliament-specific within-speaker effects that resist simple generalisation. However, there seems to be a stable global effect, also reproduced in two out of four parliaments, showing that opposition speeches have lower FP rates than ruling coalition speeches. This finding requires future in-depth research.

Taken together, our findings demonstrate the value of large, transformer-annotated, cross-lingual corpora and within-speaker decomposition for understanding FP production.
They further suggest that many effects reported in smaller studies may reflect domain- or language-specific phenomena rather than universal tendencies. 

\section{Future Work}

The domain restriction to parliamentary speech limits generalisation; extending this framework to conversational or broadcast data would test which effects are robust across registers. 
The gender reversal in South Slavic parliaments, and the consistent behaviour regarding power status with Serbia as a potential anomaly, both warrant targeted sociolinguistic investigation. Finally, this work has investigated the incidence of filled pauses, while there is additional potential in investigating the duration and type of filled pauses, as well as the linguistic context in which they occur.

\section{Acknowledgments}

This work was supported in part by the project ``Large Language Models for Digital Humanities'' (GC-0002), the project ``EPIC-SI - Early Parent-Child Communication in Slovenian: Corpus-based Insights'' (J6-70222), the research programme ``Language Resources and Technologies for Slovene'' (P6-0411), and the Research Infrastructure DARIAH-SI (I0-E007), all funded by the ARIS Slovenian Research and Innovation Agency.

\section{Generative AI Use Disclosure}

During the preparation of this manuscript, generative AI was used for improving the grammar and style of specific sections of the manuscript.

\bibliographystyle{IEEEtran}
\bibliography{mybib}

\end{document}